\documentclass[conference]{IEEEtran}

\IEEEoverridecommandlockouts                              

\usepackage{times} 
\usepackage[numbers]{natbib}

\usepackage{amsmath} 
\usepackage{amssymb}  

\usepackage[utf8]{inputenc}
\usepackage{booktabs}
\usepackage{multirow}
\usepackage{subcaption} 
\usepackage[skip=0pt,font=small,labelfont=bf]{caption}
\usepackage{array}
\usepackage{xspace}
\usepackage{flushend}
\usepackage{siunitx}

\usepackage{algorithm}
\usepackage[noend]{algpseudocode}
\usepackage{bm}

\usepackage{listings}
\usepackage[dvipsnames]{xcolor}
\DeclareMathOperator{\EX}{\mathbb{E}}

\algnewcommand{\LeftComment}[1]{\Statex \(\triangleright\) #1}

\usepackage[pagebackref=true,breaklinks=true,bookmarks=true,colorlinks,urlcolor=blue]{hyperref}
\usepackage[colorinlistoftodos,prependcaption,textsize=tiny]{todonotes}

\newcommand{\structdiffusion}{\emph{StructDiffusion}}
 
\begin{document}
\title{StructDiffusion: Language-Guided Creation \\ of Physically-Valid Structures using Unseen Objects}


\author{Weiyu Liu$^1$, Yilun Du$^2$, Tucker Hermans$^{3}$, Sonia Chernova$^1$, Chris Paxton$^4$ \thanks{
$^1$Georgia Tech, $^2$MIT, $^3$University of Utah and NVIDIA, $^4$Meta AI
}}


\setcounter{figure}{1}
\makeatletter
\let\@oldmaketitle\@maketitle
\renewcommand{\@maketitle}{\@oldmaketitle
  \begin{center}
    \includegraphics[width=\linewidth]{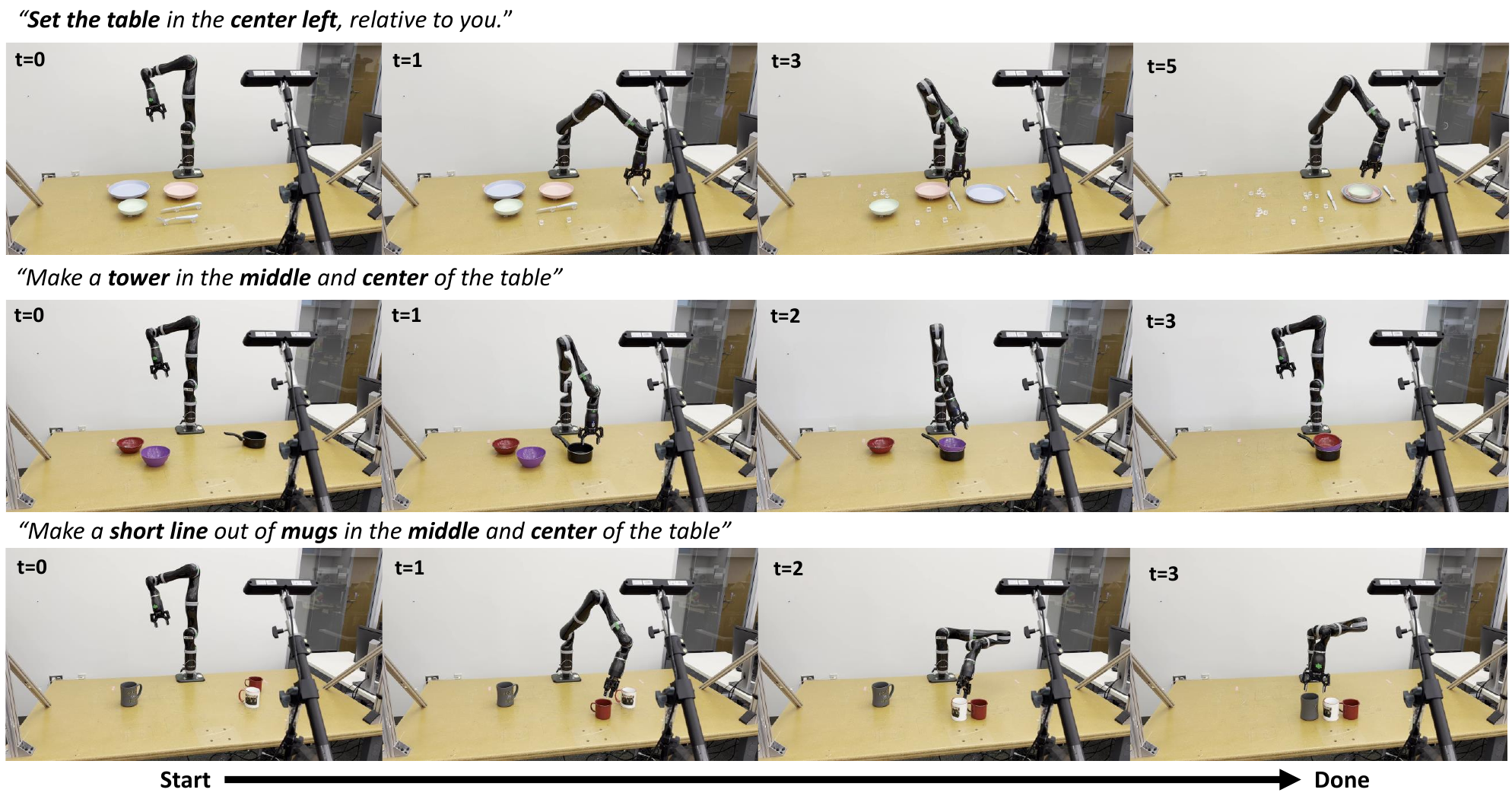}
  \end{center}
  \footnotesize{\textbf{Fig.~\thefigure:\label{fig:intro-example}}~
  Real-world rearrangement with unseen objects, given a language instruction. We use StructDiffusion to predict possible goals that satisfy physical constraints such as avoiding collisions between objects. At the core of StructDiffusion is an object-centric multimodal transformer backbone combined with a diffusion model, capable of sampling diverse high-level motion goals for language-guided rearrangement.
  }
 \vspace{-12pt}
  \medskip}
\makeatother

\maketitle


\begin{abstract}
  Robots operating in human environments must be able to rearrange objects into semantically-meaningful configurations, even if these objects are previously unseen. In this work, we focus on the problem of building physically-valid structures without step-by-step instructions. 
  We propose StructDiffusion, which combines a diffusion model and an object-centric transformer to construct structures given partial-view point clouds and high-level language goals, such as ``set the table''. 
  Our method can perform multiple challenging language-conditioned multi-step 3D planning tasks using one model.
  StructDiffusion even improves the success rate of assembling physically-valid structures out of unseen objects by on average $16\%$ over an existing multi-modal transformer model trained on specific structures.
  We show experiments on held-out objects in both simulation and on real-world rearrangement tasks.
  Importantly, we show how integrating both a diffusion model and a collision-discriminator model allows for improved generalization over other methods when rearranging previously-unseen objects. 
  For videos and additional results, see our website: \url{https://structdiffusion.github.io/}.
  
\end{abstract}

\section{Introduction}\label{sec:intro}
Structures are everywhere in the real world: shelves are stocked, tables are set, furniture is assembled. For robots to be successful assistants and collaborators, they must understand object structures and build these structures based on human commands.
In this work, we predict how previously-unseen objects should be rearranged in order to realize language instructions, given only partial-view point cloud of a scene.
Solving this task requires models that can reason about different constraints over where objects should be at once (e.g., object geometry, language-driven task semantics, physics) and generate solutions that respect all these constraints.

Assume a robot is given a language instruction such as ``set the table.'' First, the objects must in the correct relative positions in order to satisfy desired spatial and semantic relations: utensils on the sides, plate in the center, for example. Second, arrangements must be physically valid: even if the plate is bigger than the robot saw in training data, it should not collide with the utensils. These constraints correspond to two overlapping, but not identical, sets of possible solutions where a robot can place objects.

One potential approach for this task is to train a language-conditioned multi-modal transformer to directly predict the best sequence of actions the robot should take~\cite{jiang2022vima,shridhar2022cliport,shridhar2022perceiver,liu2022structformer,guhur2022instruction}. Such models have achieved impressive results on pick-and-place tasks for known objects~\cite{jiang2022vima,shridhar2022perceiver}; however, for structure generation of unseen objects, regressing to a single solution can often create problems when there are multiple, potentially conflicting, constraints (e.g., place objects ``tightly'' but avoid collisions)~\cite{liu2022structformer}. 
By contrast, planning-inspired approaches work by generating and refining a distribution over estimated poses~\cite{paxton2021predicting}, most notably diffusion models~\cite{janner2022diffuser,huang2023diffusion}. Such diffusion models have shown applications not just to image generation~\cite{rombach2022high,sohl2015deep,ho2020denoising,kapelyukh2022dall}, but to motion planning and pose estimation~\cite{huang2023diffusion,janner2022diffuser}, and to rearrangement~\cite{kapelyukh2022dall}, achieving better results than previous policies due to their ability to more accurately capture the space of potential solutions.
We hypothesize that by iteratively refining predicted goals from a diffusion model, subject to learned constraints, we can similarly find better solutions for language-conditioned structure creation.
In our approach, called \structdiffusion{}, we first use unknown object instance segmentation to break up our scene into objects, as per prior work~\cite{paxton2021predicting,singh2022progprompt,goyal2022ifor,qureshi2021nerp}. 
Then, we use a multi-modal transformer to combine both word tokens and object encodings from Point Cloud Transformer~\cite{guo2021pct} in order to make 6-DoF goal pose predictions.
These predictions are both refined iteratively via diffusion and the best goal is selected with a ``discriminator'' model that is trained to recognize unrealistic samples. Because goals are predicted relative to the observed partial-view point clouds of the objects, it's possible to get strong results without any object models or object foreknowledge.
We train two main components:
\begin{enumerate}
    \item \textbf{An object-centric, language-conditioned diffusion model}, which learns how to construct different types of multi-object structures from observations of novel objects and language instructions.
    \item \textbf{A learned discriminator}, which drastically improves performance by rejecting samples violating physical constraints.
\end{enumerate}

This approach allows \structdiffusion{} to resolve problems not possible in previous work.
Compared to an existing single-task model and a strong multi-task baseline we introduce, \structdiffusion{} achieves $16.3\%$ and $13.5\%$ higher success rates respectively.
To our knowledge, \structdiffusion{} is the first work that can build multiple structures requiring stacking and 3D reasoning, while generalizing over different unseen objects given high-level language instructions.
We believe that this approach will allow for language-guided control of robots, to be applied to more complex tasks and environments.

\section{Related Work}\label{sec:related_work}
Language-conditioned manipulation is a fast-growing area of research~\cite{shridhar2022cliport,lynch2020learning,liu2022structformer,shridhar2022perceiver,mees2022hulc,lynch2022interactive,brohan2022rt}. Recently. Say-Can~\cite{ahn2022can} showed how a large language model (LLM) can be used to sequence robot skills to respond to a wide range of natural-language queries. This work has been extended to use a map and object-centric representation of the world~\cite{chen2022open}. As this line of work leverages LLMs for reasoning, they use a purely language-based version of the world and require efforts in prompt engineering~\cite{singh2022progprompt,liang2022code,ahn2022can}. 
Another thread of work looks more closely at learning language-conditioned skills, and has achieved impressive results on a variety of pick-and-place and articulated object manipulation tasks~\cite{shridhar2022cliport,mees2022hulc,shridhar2022perceiver}; however, these approaches yet to demonstrate whether the skills can be sequenced to create more complex structures we study here. It's worth noting that several of these works have used an object-centric representation of the world~\cite{jiang2022vima,chen2022open,liu2022structformer}, where objects are first segmented or detected and then encoded separately.
For example, VIMA~\cite{jiang2022vima} used encoded object patches as input to a multimodal transformer.
These works, however, do not look specifically at generating \textit{physically realistic} structures: in our experiments, we show how these direct-regression-first approaches do not generate the same quality of structures, and that, in particular, simply predicting the best placement poses or actions will lead to more failures.

Our work is related to planning with unknown objects. Simeonov et al.~\cite{simeonov2020long} propose a planning framework for rigid objects, but they do not study structures with complex dependencies. Curtis et al.~\cite{curtis2022long} investigate task and motion planning with unknown objects, which relies on a similar segmentation and grasping pipeline we use, but does not look at learning goals, instead assume access to a set of predicates (e.g., Red, on) which can be evaluated at planning time.

Furthermore, there is a set of works which learn object-object relations for planning~\cite{paxton2021predicting,bobu2022learning,yuan2021sornet,simeonov2022relational}, which is relevant to our method's refinement process. Many of these do so explicitly. In particular, learning object skill preconditions is very useful for sequential manipulation, so some works look at predicting relationships in this context~\cite{sharma2020relational,yuan2021sornet,migimatsu2022grounding}. For example, SORNet~\cite{yuan2021sornet} learns to predict relations between objects given a canonical image view of the objects; similarly a predictive model from image inputs is learned for capturing relationships~\cite{migimatsu2022grounding}. These object-object relations are an important part of planning sequential manipulation as per \structdiffusion{}, but we pay attention to implicitly classifying object relationships as a whole. 

Finally, a set of recent works have explored how diffusion models may be applied to robotics \cite{janner2022diffuser,urain2022se3dif, huang2023diffusion, kapelyukh2022dall, ajay2022conditional}. In \cite{janner2022diffuser, urain2022se3dif, ajay2022conditional}, the diffusion model is used to parameterize the motion planning procedure, while our approach focuses on generating rearrangement goals for manipulation. Furthermore, these works require known object models and are not conditioned on flexible language goals.
Similar to our work, in DALL-E-Bot~\cite{kapelyukh2022dall},  DALL-E is used to generate a goal image for an arrangement of multiple objects, from which object matching is used to obtain an rearrangement of objects. However, this approach assumes that underlying state of world can be captured using an overhead image, 
as a generated image is used to parameterize the final goal and a motion plan. In our case, we directly generate object placement poses given point cloud observations, and are thus unrestricted by which position or angle the image is generated from.


\section{Preliminaries}\label{sec:transformer}
\begin{figure*}[ht]
\includegraphics[width=1.0\textwidth]{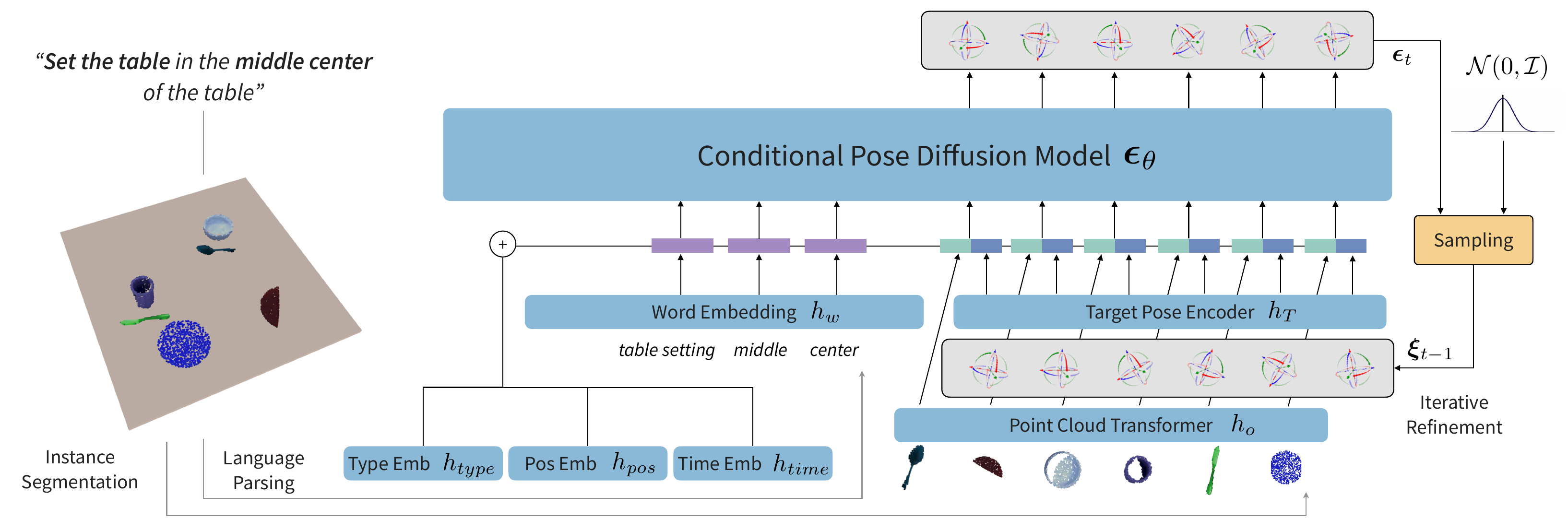}
\caption{Overview of the object-centric diffusion model. We combine a diffusion model with an object-centric multimodal transformer to iteratively reason about both 3D object embeddings and task specification in language, and to predict goal poses of objects.}
\label{fig-arch}
\vspace{-.4cm}
\end{figure*}

We provide background information on diffusion models~\cite{sohl2015deep,ho2020denoising} and transformers~\cite{vaswani2017attention}. These two neural network structures provide core components for our approach.

\subsection{Diffusion Models}
\label{sec:prelim-diffusion}

Denoising Diffusion Models are a class of generative models~\cite{sohl2015deep,ho2020denoising}. Given a sample $x \sim q(x_0)$ from the data distribution. The \textit{forward} diffusion process is a Markov chain that creates latent variables $x_1,...,x_T$ by gradually adding Gaussian noise to the sample:
\begin{align*}q(x_t|x_{t-1}) =     \mathcal{N}(x_t; \sqrt{1-\beta_t}x_{t-1}, \beta_t\mathcal{I})\end{align*}
Here $\beta_t$ follows a fixed variance schedule such that the variance at each step is small and the total noise added to the original sample in the chain is large. These two conditions allows sampling $x_0 \sim p_\theta(x_0)$ from a \textit{reverse} process that starts with a Gaussian noise $x_T$ and follows a learned Gaussian posterior
\begin{align*}
p_\theta(x_{t-1}|x_t) \sim \mathcal{N}(x_{t-1}; \mu_\theta(x_t, t), \Sigma_\theta(x_t, t))
\end{align*}
In this work, we adopt the simplified model introduced in \cite{ho2020denoising} that fixes the covariance $\Sigma_\theta(x_t, t)$ to an untrained time-dependent constant and reparameterize the mean $\mu_\theta(x_t, t)$ with a noise term $\epsilon_t$. Diffusion models can be trained to minimize the variational lower bound on the negative log-likelihood $\EX[-\log p_\theta(x_0)]$. A simplied training objective with the reparameterized mean can be derived as:
\begin{align*}
    L_\text{simple} = \EX_{t \sim [1, T], x_0 \sim q(x_0), \epsilon \sim \mathcal{N}(0, \mathcal{I})}[||\epsilon - \epsilon_\theta(x_t, t)||^2]
\end{align*}

\subsection{Transformers}
\label{sec:prelim-transformers}
Transformers were proposed in \cite{vaswani2017attention} for modeling sequential data. At the heart of the Transformer architecture is the scaled dot-product attention function, which allows elements in a sequence to attend to other elements. Specifically, an attention function takes in an input sequence $\{x_1,...,x_n\}$ and outputs a sequence of the same length $\{y_1,...,y_n\}$. Each input $x_i$ is linearly projected to a query $q_i$, key $k_i$, and value $v_i$. The output $y_i$ is computed as a weighted sum of the values, where the weight assigned to each value is based on the compatibility of the query with the corresponding key. 
In this work, we use the encoder layers in the original transformer architecture. Each encoder layer includes an attention layer and a position-wise fully connected feed forward network. With the use of attention mask, the encoder layer can process sequences with different lengths.

\section{\structdiffusion{} for Object Rearrangement}\label{sec:approach}

\begin{figure*}[bt]
\centering
\includegraphics[width=1.0\textwidth]{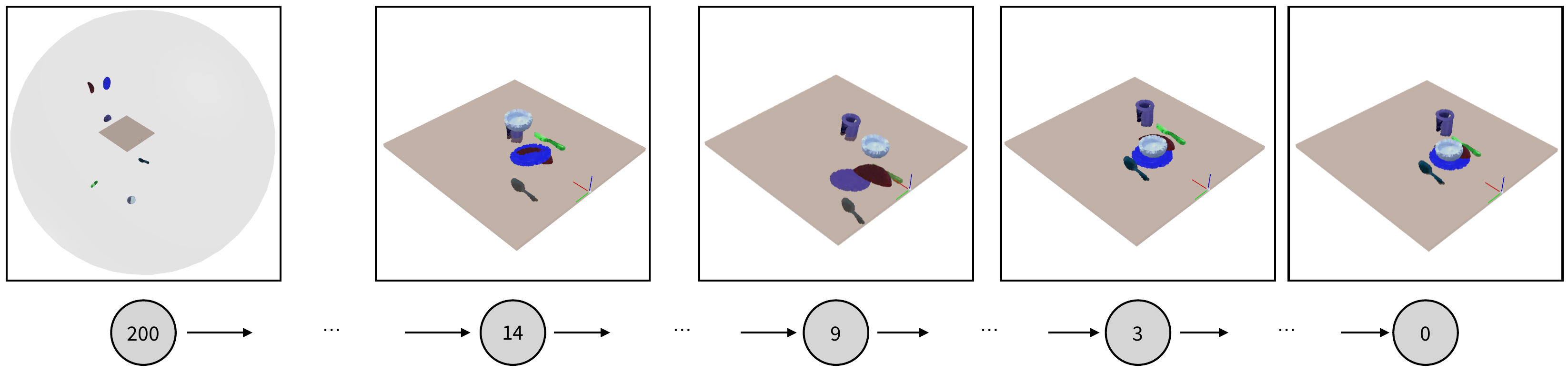}
\caption{We model goal generation for semantic rearrangement as a diffusion process. For each sample, we start from the last step of the reverse diffusion process, which places objects randomly in space, and jointly predict goal poses for all objects in the scene. This formulation allows our model to reason about object-object interactions in a generalizable way, which outperforms simply predicting goal poses from multi-modal inputs.}
\label{fig:reverse-diffusion}
\vspace{-.5cm}
\end{figure*}

Given a single view of an initial scene containing objects $\{o_1,...,o_N\}$ and a language specification containing word tokens $\{w_1,...,w_M\}$, our goal is to rearrange the objects into a goal configuration that satisfies the language goal.
We assume the objects are rigid and we are given a partial-view point cloud of the scene with segment labels for points to identify the objects. We can extract the initial poses of the objects $\{\xi^{pc}_1,...,\xi^{pc}_N\}$ from the segmented object point clouds $\{x_1,...,x_N\}$ by setting the rotation to zero and the position to the centroid of each object point cloud in the world frame. To rearrange the objects, the robot needs to move the objects to their respective goal poses $\xi^{goal}_i$. 

In this work, our robot can execute pick and place actions. For each object, we can sample a set of stable grasps $\mathcal{G} = \{g_1,...,g_M\}$. Given a target pose for object $\xi^{goal}_i$ and a stable grasp $g_j$, the robot can move its end effector to  $\xi^{ee}_i=\xi^{goal}_i (\xi^{pc}_i)^{-1} g_j$ to place the object at the goal pose. We only use pick and place actions in our setup to simplify the problem. However, our object-centric actions can be integrated with sampling-based TAMP solutions \cite{curtis2022long} to also leverage other motion primitives, such as pushing and regrasping, to reach the goal poses predicted by our system. 

Below, we describe our approach for sampling goal poses for objects based on partial point clouds, given a language goal. Our framework combines a generator based on a diffusion model and a learned discriminator that filters out invalid samples. As shown in Fig.~\ref{fig-arch}, our diffusion model is integrated with a transformer model that maintains an individual attention stream for each object. This object-centric approach allows us to focus on learning the interactions between objects based on their geometric features, as well as the grounding of abstract concepts on spatio-semantic relations between objects (e.g., large, circle, top). The discriminator model operates on the imagined scene after transforming the objects to their predicted goal poses to further reject invalid samples.


\subsection{Encoders}
We leverage modality-specific encoders to convert the multimodal inputs to latent tokens that are later processed by the transformer network.

\textbf{Object encoder.}
Given the segmented point cloud $x_i$ of an object $o_i$, we learn an encoder $h_o(x_i)$, in order to obtain the latent representation of the object.
This is based on Point Cloud Transformer (PCT)~\cite{guo2021pct},
which has been shown to be effective at shape classification and part segmentation. We process the centered point cloud with PCT and learn a separate multilayer perceptron (MLP) to encode the mean position of the original point cloud. Encodings from the two networks are concatenated to give $h_o(x_i)$.
We rely on this latent representation of objects for semantic, geometric, and spatial reasoning.

\textbf{Language.} To map the language goal to its latent representation, we map each unique word token from the language instructions separately to an embedding with a learned mapping $h_w(w_i)$. This method helps establish a fine-grained correspondence between each part of the language specification and the respective constraint on the generated structure. We further investigate encoding whole sentences using a sentence transformer in Appendix \ref{sec:sentence-model}.

\textbf{Diffusion encodings.} Since the goal poses of objects are iteratively optimized by the diffusion model and need to feed back to the model, we use a MLP to encode the goal poses of the objects $h_T(\xi^{goal}_i)$. To compute the time-dependent Gaussian posterior for reverse diffusion, we combine a latent code for $t$ in the feature channel by learning a time embedding $h_{time}(t)$.

\textbf{Positional encoding.} To differentiate the multimodal data, we use a learned position embedding $h_{pos}(i)$ to indicate the position of the words and objects in input sequences and a learned type embedding $h_{type}(\upsilon_i)$ to differentiate object point clouds ($\upsilon_i=1$) and word tokens ($\upsilon_i=0$). 



\subsection{Conditional Pose Diffusion Model}
Combining a diffusion model and an object-centric transformer,
\structdiffusion{} can sample diverse yet realistic 
object structures while accounting for the complex constraints imposed by the object geometry and language goal. The conditional diffusion model predicts the goal poses for the objects $\bm{\xi}_0 = \{\xi_i\}^N_i$ starting from the last time step of the reverse diffusion process $\bm{\xi}_T \sim \mathcal{N}(0, \mathcal{I})$, as illustrated in Fig.~\ref{fig:reverse-diffusion}. We use the bold symbol here because we jointly optimize the poses of all objects.

\textbf{Object-centric transformer.} Different from most existing diffusion models that directly generate goal images and do not explicitly model individual objects \cite{rombach2022high,sohl2015deep,ho2020denoising,kapelyukh2022dall}, we use the transformer model to build an object-centric representation of the scene and reason about the higher-order interactions between multiple objects. This approach allows us to account for both global constraints and local interactions between objects. Leveraging attention masks, a single transformer model can also learn to rearrange different numbers of objects. 

\textbf{Diffusion.} The use of the diffusion model helps us capture diverse structures since we are sampling from a series of Gaussian noises at different scales when going from $\bm{\xi}_T$ to our goal $\bm{\xi}_0$. The resulting samples, therefore, is diverse at different levels of granularity (e.g., different placements of the structures and different orientations of the individual objects). Diversity is also crucial when dealing with the inherent ambiguity in language instructions. For example, a \textit{large} circle of plates and a \textit{large} circle of candles impose different constraints on the proportions of the structures because the objects being arranged have different sizes.

\medskip

Combining the advantages of the object-centric transformer and the diffusion model, we propose to model the conditional reverse process as 
\begin{align*}
    p_\theta(\bm{\xi}_0| \{x_i\}, \{w_i\}) = p(\bm{\xi}_t) \prod p_\theta(\bm{\xi}_{t-1}| \bm{\xi}_{t}, \{x_i\}, \{w_i\})
\end{align*}
The generation process depends on the point clouds of the objects and language instruction. As discussed in \ref{sec:prelim-diffusion}, we learn the time-dependent noise $\bm{\epsilon}_t$, which can be used to compute $\bm{\xi}_t$. We use the transformer as the backbone to predict the conditional noise $\bm{\epsilon}_\theta(\bm{\xi}_t, t, \{x_i\}, \{w_i\})$ for each object. We obtain the transformer input for the language part and the object part as 
\begin{align*}
    c_{i,t} & = [h_w(x_i); h_{pos}(i); h_{type}(\upsilon_i); h_{time}(t)] \\
    e_{i,t} & = [h_o(x_i); h_T(\xi^{goal}_i); h_{pos}(i); h_{type}(\upsilon_i); h_{time}(t)]
\end{align*}
\noindent where $[;]$ is the concatenation at the feature dimension. The model takes in the sequence $\{c_{1,t},..,c_{M,t}, e_{1,t}, ...,e_{N,t}\}$ and predicts $\{\epsilon_{1,t},...,\epsilon_{N,t}\}$ for the object poses. We parameterize 6-DoF pose target $\xi$ as $(t, R) \in SE(3)$. We directly predict $t \in \mathbb{R}^3$ and predict two vectors $a, b \in \mathbb{R}^3$, which are used to construct the rotation matrix $R \in SO(3)$ using a Gram–Schmidt-like process proposed in \cite{zhou2019continuity}.

\subsection{Discriminators}
In addition to the generator, we also use a learned discriminator model to further filter the predictions for realism. The discriminator works on imagined scenes, where the point clouds of objects are rigidly transformed to the respective goal poses following $x_i^{goal} = \xi^{goal}_i (\xi^{pc}_i)^{-1}x_i$. As the discriminator is trained to predict a score $s \in [0, 1]$, it can be used to rank the generated samples from the diffusion model during inference. 

Here we also have the opportunity to leverage a spatial abstraction different from the one used by the generator. The generator operates on the latent object-centric representation that are suitable to \textit{imagine} possible structures. The discriminator can directly reason about the interactions between the transformed point cloud objects at the point level. To maintain the ability to distinguish each individual object, we add a one-hot encoding to each point feature.

In our preliminary experiments, we found that this point-level collision model has more discrimination power than an object-centric model that operates on latent representations of the objects. We also investigated whether a discriminator can be integrated into the diffusion process using a technique called \textit{classifier guidance}~\cite{dhariwal2021diffusion}, but observed no significant benefits. Details are presented in Appendix~\ref{sec:classifier-guidance}.

We explore two potential discriminators.
The \textit{collision discriminator} learns to predict pairwise collisions between two objects from their partial point clouds.
Second, the \textit{structure discriminator} learns to classify the validity of the whole multi-object structure.
The structure discriminator is also language-conditioned, so it can learn structure-specific constraints. 
We found that the structure discriminator works better when it is only required to predict if local constraints are satisfied. Therefore, we normalize the scene point cloud and drop parts of the language instruction that specify global constraints such as where to place the structure on the table. 


\subsection{Planning and Inference}

\begin{algorithm}[bt]
\caption{Goal generation with \structdiffusion{}}
\label{alg:diffusion_planning}
\begin{algorithmic}[1]
\For{$t \in \text{range}(T, 1)$}
    \State $\bm{\epsilon}_t \sim \bm{\epsilon}_\theta(\bm{\xi}_t, t, \{x_i\}, \{w_i\})$
    \State $\bm{z} \sim \mathcal{N}(0, \mathcal{I})$ if $t>1$ else $\bm{z} = 0$
    \State $\bm{\xi}_{t-1} = \frac{1}{\sqrt{\beta_t}}(\bm{\xi}_t -  \frac{\beta_t}{\sum_{s=1}^t 1-\beta_t} \bm{\epsilon}_t) + \sqrt{\beta_t} \bm{z}$
\EndFor
\State Transform object points: $x_i^{goal} = \xi^{goal}_i (\xi^{pc}_i)^{-1}x_i$
\State Compute discriminator scores
\State \textbf{return} ranked $\bm{\xi_0}$
\end{algorithmic}
\end{algorithm}

In Alg.~\ref{alg:diffusion_planning}, we show how to combine the different components of our framework to sample object structures. We first initialize a batch of goal poses $\mathbb{R}^{\in B \times N \times (3+3+3)}$ with random noise. We use batch operation on a GPU to efficiently perform diffusion and transform point clouds of multiple objects for different samples. For the discriminators, we also generate point clouds combining all objects after the diffusion process and score them in batches. The ranked samples are returned. Each sample corresponds to a physically and semantically valid multi-object structure that can be used by other components of the manipulation pipeline for planning.

\subsection{Training Details}
We train the diffusion model and the discriminators separately. For the diffusion model, we use the dataset from \cite{liu2022structformer} containing high-level language instructions, segmented object point clouds, and sequences of rearrangement actions. We extract the goal poses from the rearrangement actions. We train a single model for all structures where the number of examples for different classes of structures are balanced. We use a batch size of 128 and train the diffusion model on an RTX3090 GPU for about 12 hours. To train the collision discriminator, we randomly sample $100,000$ pairwise configurations for objects from the dataset. For the structure discriminator, we generate negative examples by randomly perturbing the ground truth target poses $\xi^{goal}_i$.
For each negative example, we randomly select a subset of objects to perturb so that the negative examples have different numbers of objects out of place. We augment the data by using point clouds from different time steps of the rearrangement sequence to create the imagined scenes as they usually occlude different objects.





\section{Simulation Experiments}\label{sec:experiments}

We first evaluate our method in simulation by comparing it to existing methods and also a collection of strong baselines we introduce for language-guided multi-object rearrangement. 

\subsection{Baselines}
We compare our approach against the following baselines:

\smallskip \noindent 
\textbf{StructFormer}: This prior method uses a multimodal transformer network to generate multi-object structures based on segmented object point clouds and language instructions \cite{liu2022structformer}.  The transformer network autoregressively predicts the goal poses of each object.  We follow the original work to train a separate model for each class of structure. 

\smallskip \noindent 
\textbf{Conditional Variational Autoencoder (CVAE)}: 
CVAEs have been used to capture different modes for multi-task learning and language-conditioned manipulation~\cite{lynch2020learning,mees2022hulc}.
Our CVAE baseline uses the object-centric transformer backbone as a strong baseline for semantic rearrangement.  
    To prevent the latent variable from being ignored when combining the transformer with CVAE, the transformer network predicts the object goal poses in a single forward pass (i.e., not autoregressively).  A single model is trained for four classes of structures. 

\smallskip \noindent \textbf{Optimization with Learned Discriminator}: This baseline iteratively optimizes the goal poses of objects with the \textit{structure discriminator} that is trained to classify valid rearranged scenes and invalid ones. This general approach has been used extensively for learning language-conditioned manipulation from offline data~\cite{nair2022learning}, grasping~\cite{murali20206,lu2020multifingered}, and predicting stable placements of objects~\cite{paxton2021predicting}, but not for language-conditioned multi-object rearrangement. We use the cross-entropy method for optimization~\cite{chen2021fastrack},
and only optimize the object poses and not the structure pose to simplify the optimization problem. We initialize samples from the baseline generative models because initializing with random values does not lead to meaningful performance.


\begin{figure}[bt]
\includegraphics[width=1.0\columnwidth]{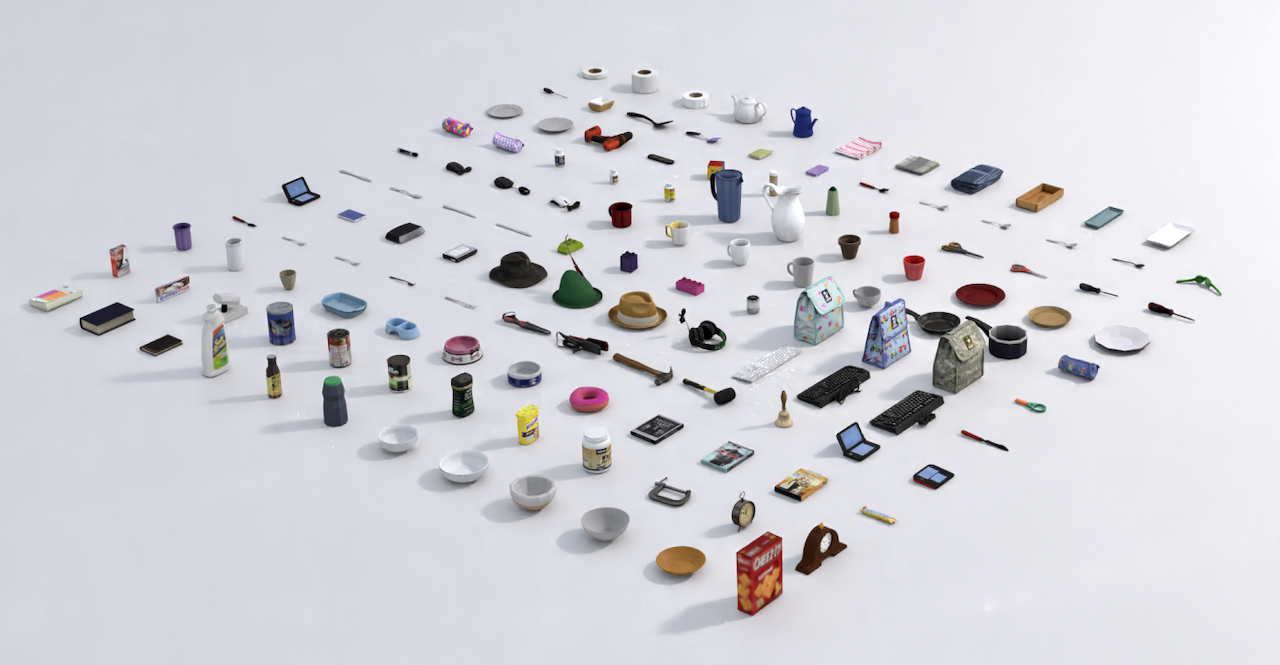}
\centering
\caption{
Testing objects from Google Scanned Objects~\cite{downs2022google}, ReplicaCAD dataset~\cite{szot2021habitat}, and the YCB Object Set~\cite{calli2015ycb}. The test object dataset contains a wide range of textured objects belonging to various classes. None of these objects appear in the training data.
}
\label{fig:objects}
\vspace{-.5cm}
\end{figure}

\subsection{Experimental Setup}
We evaluate all models in the PyBullet physics simulator~\cite{coumans2017pybullet}. Point cloud observations are rendered with NViSII~\cite{morrical2021nvisii}. We test on novel object models from both known and unknown categories as our goal is to transfer the model learned in simulation directly to real-world objects. Fig.~\ref{fig:objects} shows the testing objects, which are collected from Google Scanned Objects~\cite{downs2022google}, ReplicaCAD dataset~\cite{szot2021habitat}, and the YCB object Set~\cite{calli2015ycb}. To generate the test scenes, we use the same data collection pipeline that is used to collect ground truth data from prior work~\cite{liu2022structformer}. This ensures that a valid rearrangement can be found for each scene. The set of objects and the language goal for each scene are randomly sampled. Distractor objects are randomly placed in the scene to simulate occlusions. 

We report success rate for the rearrangements. To isolate the pose prediction problem from other components of the system (e.g., grasp sampling and motion planning), we directly place objects $3$cm above the the predicted target poses. We checks whether the rearrangement is physically valid by running the simulation loop after placing each object. We check possible collisions and intersections between objects using approximate convex decompositions of the 3D object models. We also implement model-based classifiers to evaluate whether the rearrangement satisfy the language goal. For example, we check whether the objects are in a line using the centroids of the models. 
A rearrangement is considered as successful if the placements of objects are not preempted due to physics-related failures and the goal scene satisfies all semantic constraints determined by the given language goal. On average, there are 5 constraints for different types of structures.

\begin{figure}[bt]
\includegraphics[width=1.0\columnwidth]{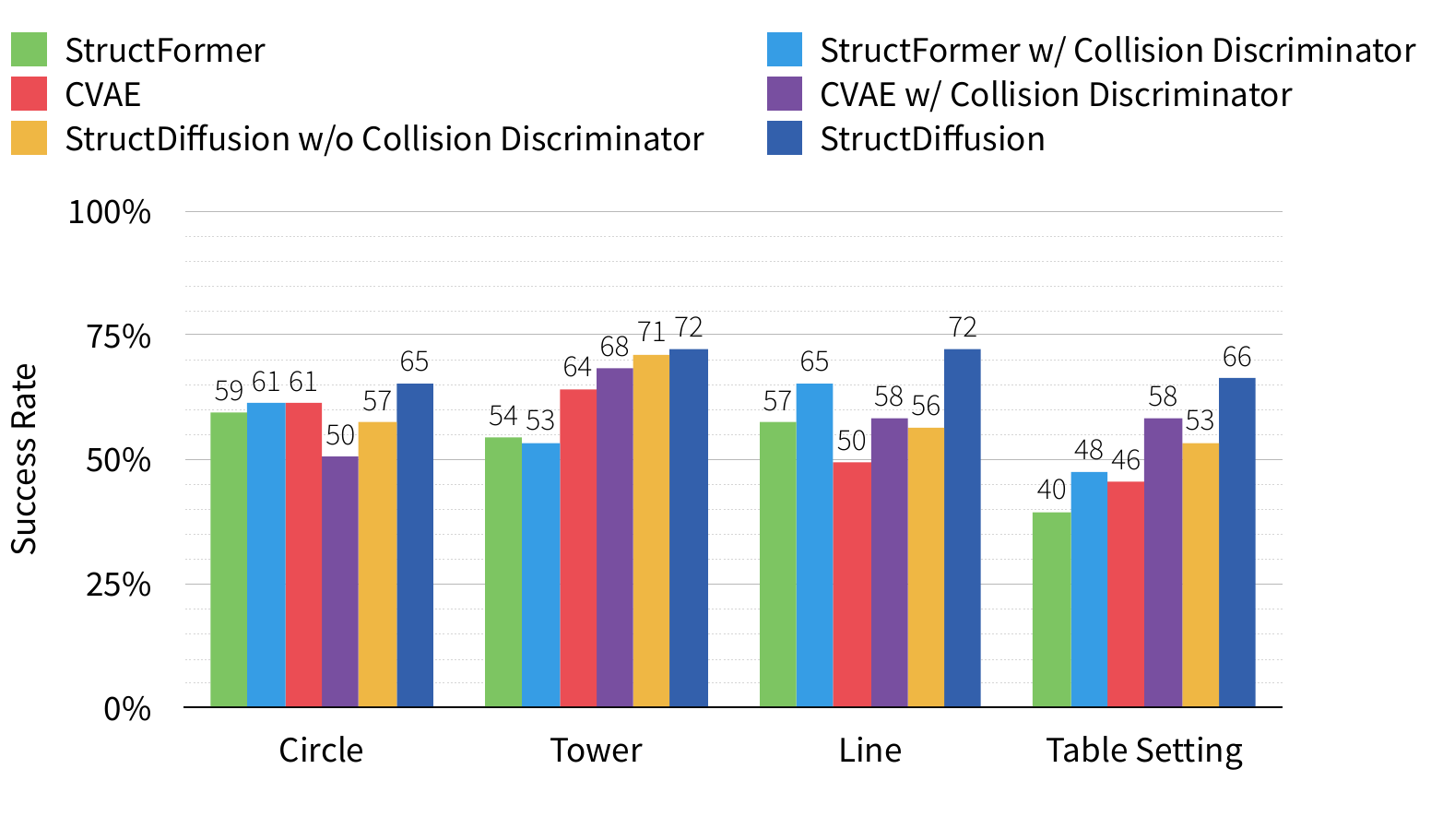}
\centering
\caption{Success rates for four different classes of structures on held-out objects. Models are evaluated in a physics simulator using unseen objects. A rearrangement is successful only if all objects are placed in physically valid poses and the rearranged scene satisfies the language goal. Compared to StructFormer~\cite{liu2022structformer}, the model previously proposed for semantic rearrangement, \structdiffusion~obtains a 16\% average improvement in success rate.}
\label{fig:result-generative}
\end{figure}

\begin{figure}[bt]
\includegraphics[width=1.0\columnwidth]{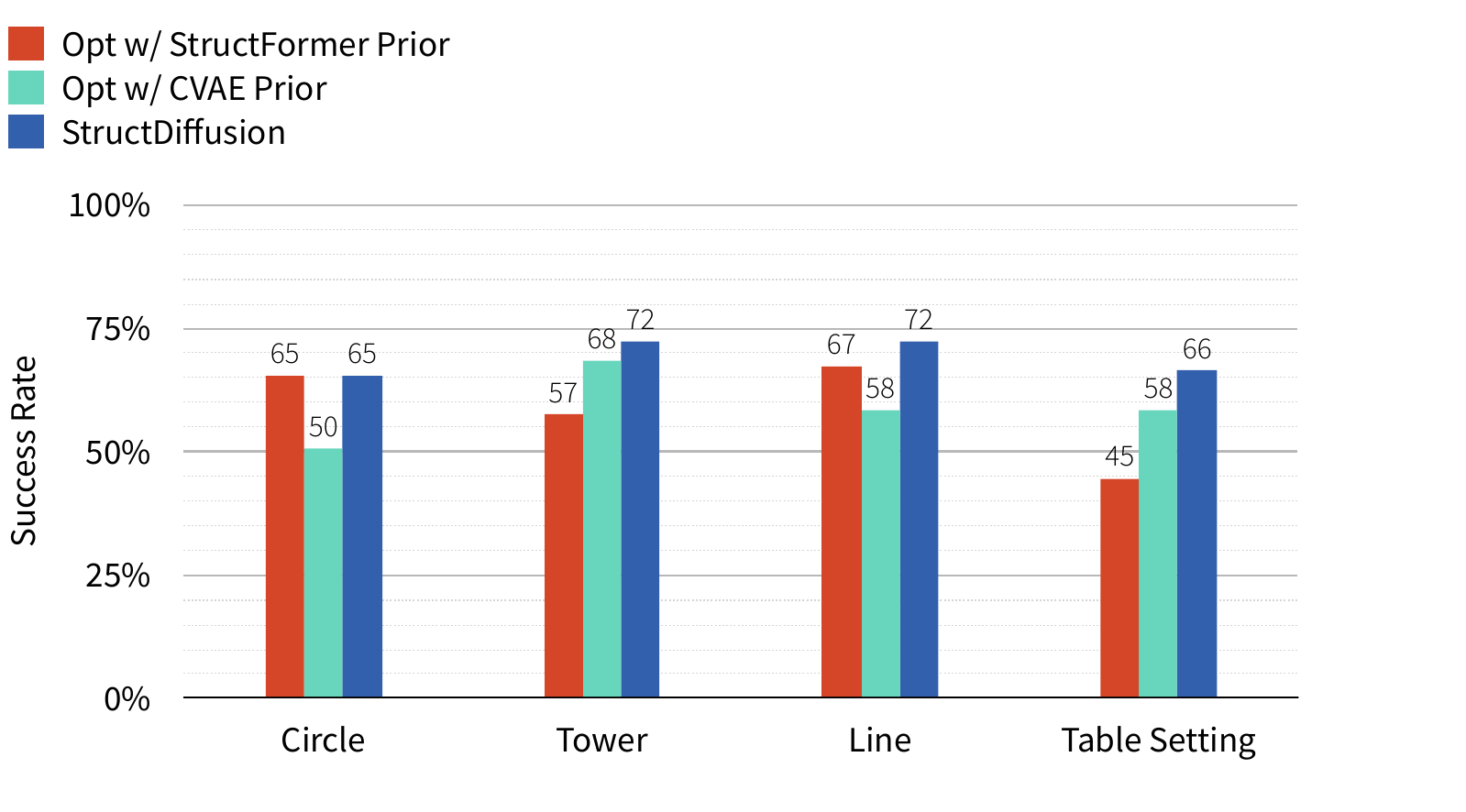}
\centering
\caption{Comparing \structdiffusion{} with other iterative methods. The two baselines initialize samples of target object poses using either the StructFormer or the CVAE model. The predicted scores of a learned discriminator is then used to guide iterative optimization of the samples. In comparison, \structdiffusion{} directly predicts the noises $\epsilon_t$ that need to be removed from the samples at each step.}
\vspace{-.5cm}
\label{fig:result-iterative}
\end{figure}

\begin{figure*}[bt]
\centering
\includegraphics[width=1.0\textwidth]{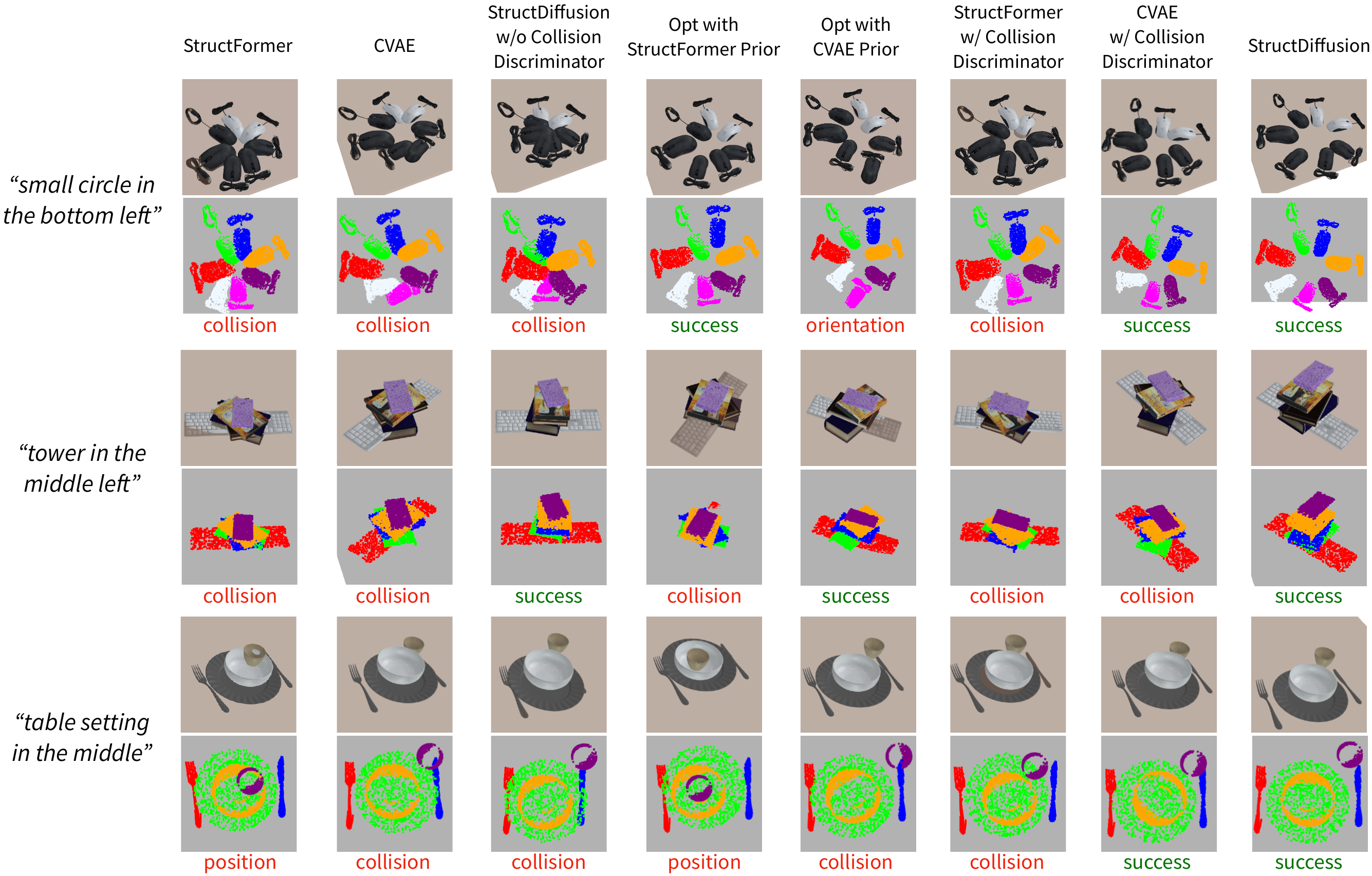}
\caption{Comparison between \structdiffusion{} and the baselines on partial views of held-out objects, given language commands from four different categories. \structdiffusion{} is better at resolving constraints involving contact and precise arrangement of objects, avoiding collisions and creating physically realistic placements. The labels indicate whether the structures can be successfully built in the simulation environment and also satisfy the language goal.}
\label{fig:pc-rearrangement-comparison}
\vspace{-.5cm}
\end{figure*}

\subsection{Comparison with Other Generative Models}
In Fig.~\ref{fig:result-generative}, we compare with other generative models and gain insights into the generator-discriminator design of our model. We see that our complete model, \structdiffusion{}, significantly outperformed all baselines on all structure types. The improvement was most significant for structures that required precise placements of objects and modeling contacts between objects. The generator-discriminator design was necessary because the diffusion model alone still generated invalid samples, especially for the line structures. The performance difference between \structdiffusion{} and the ablated model that does not use discriminator supports that our model can leverage the complementary strengths of object-centric representation and scene-level representation that preserves the point-to-point interactions.  In our preliminary experiments, we observed that using the collision discriminator is enough to create high-quality samples; however, this is not the case for the iterative methods discussed in the following section.

Although applying the collision discriminator also improved the performance of StructFormer and the CVAE, our diffusion model benefited the most from the addition. We attribute this difference to the different diversities of samples from these three classes of generative models. The autoregressive transformer underlying StructFormer does not explicitly model uncertainty, therefore leading to similar samples for each scene. The single source of stochasticity from the latent variable of the CVAE model is also not enough. As the diffusion model incorporates uncertainties at different scales, it has the ability to generate different classes of structures but also generate hypotheses of object placements given only partial, and even heavily occluded, point clouds of objects. We provide a qualitative comparison in Fig.~\ref{fig:pc-rearrangement-comparison}. We further break down the failure cases based on structure types and methods to support the insights discussed above in Appendix~\ref{sec:failure-mode-analysis}.

\subsection{Comparison with Other Iterative Methods}
In Fig.~\ref{fig:result-iterative}, we compare \structdiffusion{} with other optimization-based baselines that can take advantage of the additional computational time to iteratively refine the prediction.
The result shows that \structdiffusion{} outperformed the other two baselines. Even though we observed strong performance when applying an optimization-based method~\cite{paxton2021predicting} to other manipulation tasks, we do not see significant benefit in our task. Looking more closely, we observe that the challenging cases that are not yet solved by the non-iterative variants are cases where the placements of objects are highly related. 
In these cases, the guidance from the discriminator can be ambiguous and leads to local minimal without reaching valid solutions. We hypothesize that leveraging guidance at different scales is necessary, as studied in a recent work that directly learns to predict scores (i.e., gradients) at different scales for 2D object rearrangement~\cite{wu2022targf}. 


\section{Real World Experiments}\label{sec:robot}



\begin{figure*}[bt]
\includegraphics[width=\textwidth]{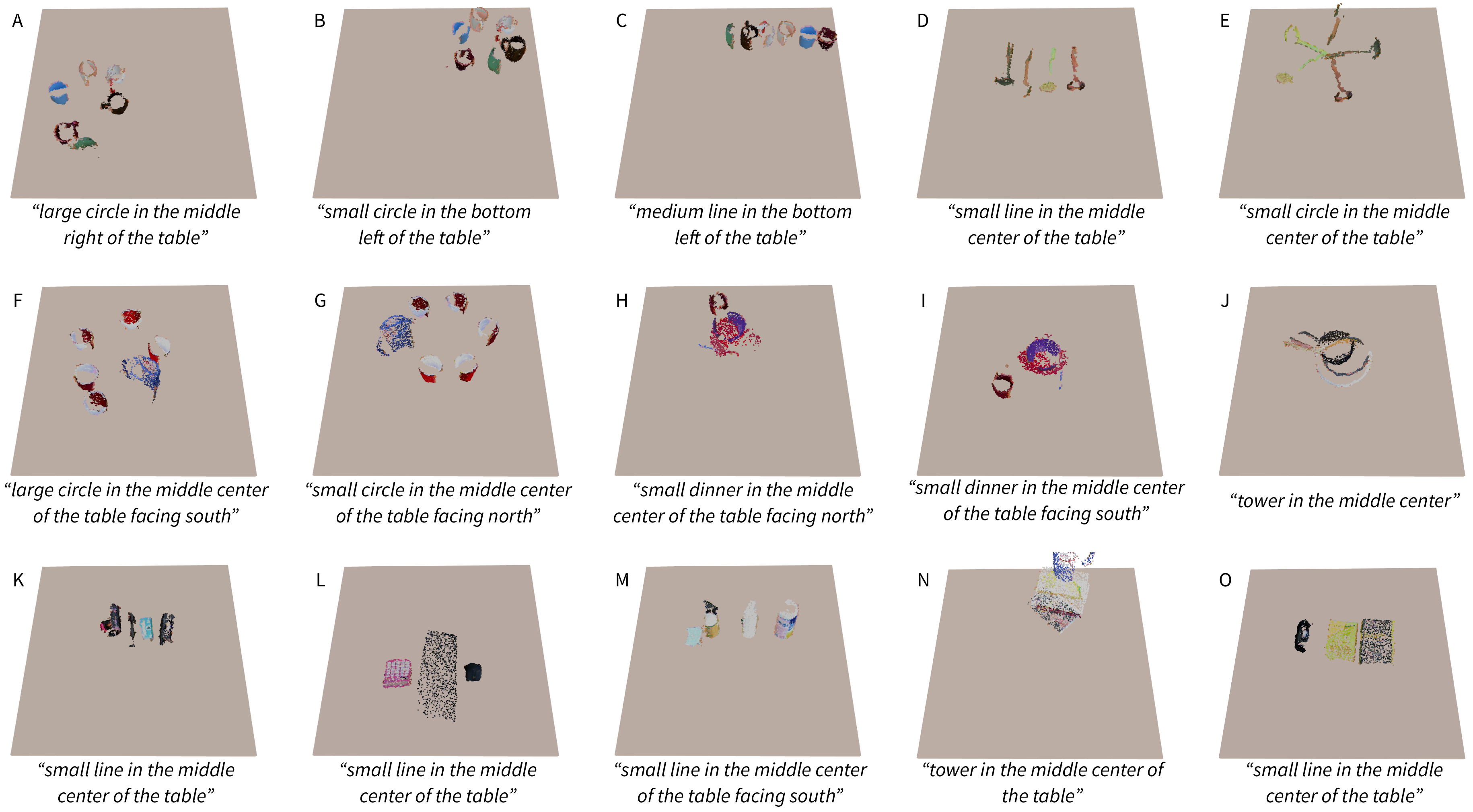}
\caption{Examples of predicted structures for real-world objects. We can predict structures from raw point clouds for a wide range of language instructions fitting into four different broad classes. Note the point clouds are incomplete because they were captured from a single camera.}
\label{fig:pc-rearrangement-examples}
\end{figure*}


In this section, we report on real-world experiments, testing structure assembly on a real robot.

\subsection{Perception and Hardware}

We deployed our system on a 7-DoF JACO arm with an Asus Xtion RGB-D Camera. We obtained segmented object point clouds by identifying clusters-of-interest through table surface detection and Euclidean distance clustering, using the Point Cloud Library~\cite{rusu20113d}. For most objects, we calculated antipodal grasps over each object point cloud~\cite{ten2018using}, which are then ordered and executed using pairwise ranking~\cite{kent2018adaptive}. For objects involved in the table setting arrangements (e.g., forks and plates), we sampled grasps using the Contact-GraspNet~\cite{sundermeyer2021contact} because antipodal grasp sampling failed to generate successful grasps for these objects. We used RRT-Connect~\cite{kuffner2000rrt} for motion planning. We released each object $3$cm above the predicated target pose for placement.


\subsection{Predictions for Real-World Objects}

We first show qualitative examples of the predicted structures for real-world objects in Fig.~\ref{fig:pc-rearrangement-examples}. These examples are created by rigidly transforming the segmented object point clouds from an initial scene with the target poses of the highest ranked structure. Even though our model is trained only on simulation data, it can be directly used to generate semantically diverse and physically valid structures for real-world objects. Our model can generate different variations of the same structure type, as shown in \textit{(A, B)}. The same set of objects can be arranged into completely different classes of structures conditioning on the language, as shown in \textit{(A, C)} and \textit{(D, E)}. Besides changing the positions and sizes of the structures, the orientations of the structures can also be specified in language \textit{(F, G)} and \textit{(H, I)}. Note that even though table settings in the training data are only aligned horizontally as shown in \textit{I}, the use of language and training on other orientation-specific structures enable compositional generalization to a new orientation shown in \textit{H}. Finally, we see non-symmetrical objects (e.g., mugs, knifes, and spatulas) are correctly aligned in \textit{B, D, E, J, H}.

\begin{table}[t]
  \centering
  \caption{Robot experiments with real-world objects. We perform each of the task multiple times with different objects and initial placements. We show the number of times that valid grasp and motion plans are found and that the plans are executed successfully by the robot.}
  \resizebox{0.49\textwidth}{!}{\begin{tabular}{lccc}
    \toprule
    \multirow{2}{*}{Object Categories (\#Trials)} & \multirow{2}{*}{Structure} & Grasp and & \multirow{2}{*}{Placement} \\  
    &&Motion Planning& \\
    \midrule
    Bowl, Pan (3) & Small Circle & 3 & 3\\
    Bowl, Pan (3) & Tower  & 3 & 2\\
    Bowl, Pan (3) & Small line & 2 & 2\\
    Plate, Spoon, Fork, Knife, Cup (12) & Table Setting &5 &4 \\
    \midrule
    \multicolumn{2}{c}{Overall Success Rate} & 61.9\% & 52.4\% \\
    \bottomrule
  \end{tabular}}
  \label{tab:robot_result}
\end{table}

\subsection{Rearrangement with Pick and Place}

To reliably rearrange multiple objects, we combined \structdiffusion{} with grasp and motion planning. We performed nested search to find the target structure to execute. Specifically, we iterate through the generated and ranked structures. For each structure, we sample a set of grasp poses for each object and compute corresponding pre-grasp, standoff, and placement poses based on the prediction. We searched for valid motion plans between these waypoints. If all motion plans have been found, we execute them on the robot.

In Table~\ref{tab:robot_result}, we show success counts and average success rate for trials with different objects and different language goals. For circle, tower, and line structure, valid motion and grasp plans can be found most of the time due to the diverse structures generated by \structdiffusion{}. For table settings, grasp planning is still challenging due to the small tolerance of the grasp regions. The limited diversity of sampled grasps also leads to no valid inverse kinematic solutions in some cases. We observed that partial point clouds due to noisy real sensor and self-occlusions for large objects led to a small number of invalid structure predictions. While planning, we make the assumption that the objects are rigidly attached to the gripper after grasping without slippage. This assumption generally did not hold in the real world and led to occasional failures. This assumption can be relaxed by predicting a post-grasp displacement, using learned models such as \cite{zhao2020towards}. We show examples of successful executions and failure cases in Appendix~\ref{sec:extra-real-world}.

\section{Conclusions}\label{sec:conclusions}
We introduced \structdiffusion{}, an approach for creating physically-valid structures using multimodal transformers and diffusion models. \structdiffusion{} operates on point cloud images of previously-unseen objects, and can create structures for a range of language instructions. 

Specifically, we compared to a wide variety of strong baselines,
including the previous state of the art~\cite{liu2022structformer} and to a conditional variational autoencoder. End-to-end policies do not perform as well, because they cannot refine placement poses that are \textit{nearly} correct. Using diffusion models for generating diverse samples and a trained discriminator to filter object goal poses significantly improves performance. 

In this work, we did not look at optimally planning. In the future, we could look at combining this approach with task and motion planning for unknown objects, as in~\cite{curtis2022long}. Additional experiments presented in Appendix~\ref{sec:sentence-model} showed that our model could easily be combined with a pretrained language model to deal with more natural sentences; therefore, we also hope to apply our work to a far wider range of structures and more complex language commands.


\bibliographystyle{IEEEtran}
\bibliography{main}

\begin{thebibliography}{10}
\providecommand{\url}[1]{#1}
\csname url@samestyle\endcsname
\providecommand{\newblock}{\relax}
\providecommand{\bibinfo}[2]{#2}
\providecommand{\BIBentrySTDinterwordspacing}{\spaceskip=0pt\relax}
\providecommand{\BIBentryALTinterwordstretchfactor}{4}
\providecommand{\BIBentryALTinterwordspacing}{\spaceskip=\fontdimen2\font plus
\BIBentryALTinterwordstretchfactor\fontdimen3\font minus
  \fontdimen4\font\relax}
\providecommand{\BIBforeignlanguage}[2]{{%
\expandafter\ifx\csname l@#1\endcsname\relax
\typeout{** WARNING: IEEEtran.bst: No hyphenation pattern has been}%
\typeout{** loaded for the language `#1'. Using the pattern for}%
\typeout{** the default language instead.}%
\else
\language=\csname l@#1\endcsname
\fi
#2}}
\providecommand{\BIBdecl}{\relax}
\BIBdecl

\bibitem{jiang2022vima}
Y.~Jiang, A.~Gupta, Z.~Zhang, G.~Wang, Y.~Dou, Y.~Chen, L.~Fei-Fei,
  A.~Anandkumar, Y.~Zhu, and L.~Fan, ``Vima: General robot manipulation with
  multimodal prompts,'' \emph{arXiv preprint arXiv:2210.03094}, 2022.

\bibitem{shridhar2022cliport}
M.~Shridhar, L.~Manuelli, and D.~Fox, ``Cliport: What and where pathways for
  robotic manipulation,'' in \emph{Conference on Robot Learning}.\hskip 1em
  plus 0.5em minus 0.4em\relax PMLR, 2022, pp. 894--906.

\bibitem{shridhar2022perceiver}
------, ``Perceiver-actor: A multi-task transformer for robotic manipulation,''
  \emph{arXiv preprint arXiv:2209.05451}, 2022.

\bibitem{liu2022structformer}
W.~Liu, C.~Paxton, T.~Hermans, and D.~Fox, ``Structformer: Learning spatial
  structure for language-guided semantic rearrangement of novel objects,'' in
  \emph{2022 International Conference on Robotics and Automation (ICRA)}.\hskip
  1em plus 0.5em minus 0.4em\relax IEEE, 2022, pp. 6322--6329.

\bibitem{guhur2022instruction}
P.-L. Guhur, S.~Chen, R.~Garcia, M.~Tapaswi, I.~Laptev, and C.~Schmid,
  ``Instruction-driven history-aware policies for robotic manipulations,''
  \emph{arXiv preprint arXiv:2209.04899}, 2022.

\bibitem{paxton2021predicting}
C.~Paxton, C.~Xie, T.~Hermans, and D.~Fox, ``Predicting stable configurations
  for semantic placement of novel objects,'' in \emph{Conference on Robot
  Learning (CoRL)}, 2021.

\bibitem{janner2022diffuser}
M.~Janner, Y.~Du, J.~Tenenbaum, and S.~Levine, ``Planning with diffusion for
  flexible behavior synthesis,'' in \emph{International Conference on Machine
  Learning}, 2022.

\bibitem{huang2023diffusion}
S.~Huang, Z.~Wang, P.~Li, B.~Jia, T.~Liu, Y.~Zhu, W.~Liang, and S.-C. Zhu,
  ``Diffusion-based generation, optimization, and planning in 3d scenes,''
  \emph{arXiv preprint arXiv:2301.06015}, 2023.

\bibitem{rombach2022high}
R.~Rombach, A.~Blattmann, D.~Lorenz, P.~Esser, and B.~Ommer, ``High-resolution
  image synthesis with latent diffusion models,'' in \emph{Proceedings of the
  IEEE/CVF Conference on Computer Vision and Pattern Recognition}, 2022, pp.
  10\,684--10\,695.

\bibitem{sohl2015deep}
J.~Sohl-Dickstein, E.~Weiss, N.~Maheswaranathan, and S.~Ganguli, ``Deep
  unsupervised learning using nonequilibrium thermodynamics,'' in
  \emph{International Conference on Machine Learning}.\hskip 1em plus 0.5em
  minus 0.4em\relax PMLR, 2015, pp. 2256--2265.

\bibitem{ho2020denoising}
J.~Ho, A.~Jain, and P.~Abbeel, ``Denoising diffusion probabilistic models,''
  \emph{Advances in Neural Information Processing Systems}, vol.~33, pp.
  6840--6851, 2020.

\bibitem{kapelyukh2022dall}
I.~Kapelyukh, V.~Vosylius, and E.~Johns, ``Dall-e-bot: Introducing web-scale
  diffusion models to robotics,'' \emph{arXiv preprint arXiv:2210.02438}, 2022.

\bibitem{singh2022progprompt}
I.~Singh, V.~Blukis, A.~Mousavian, A.~Goyal, D.~Xu, J.~Tremblay, D.~Fox,
  J.~Thomason, and A.~Garg, ``Progprompt: Generating situated robot task plans
  using large language models,'' \emph{arXiv preprint arXiv:2209.11302}, 2022.

\bibitem{goyal2022ifor}
A.~Goyal, A.~Mousavian, C.~Paxton, Y.-W. Chao, B.~Okorn, J.~Deng, and D.~Fox,
  ``Ifor: Iterative flow minimization for robotic object rearrangement,''
  \emph{arXiv preprint arXiv:2202.00732}, 2022.

\bibitem{qureshi2021nerp}
A.~Qureshi, A.~Mousavian, C.~Paxton, M.~Yip, and D.~Fox, ``Nerp: Neural
  rearrangement planning for unknown objects,'' in \emph{Proceedings of
  Robotics: Science and Systems}, 2021.

\bibitem{guo2021pct}
M.-H. Guo, J.-X. Cai, Z.-N. Liu, T.-J. Mu, R.~R. Martin, and S.-M. Hu, ``{PCT}:
  Point cloud transformer,'' \emph{Computational Visual Media}, vol.~7, no.~2,
  pp. 187--199, 2021.

\bibitem{lynch2020learning}
C.~Lynch, M.~Khansari, T.~Xiao, V.~Kumar, J.~Tompson, S.~Levine, and
  P.~Sermanet, ``Learning latent plans from play,'' in \emph{Conference on
  robot learning}.\hskip 1em plus 0.5em minus 0.4em\relax PMLR, 2020, pp.
  1113--1132.

\bibitem{mees2022hulc}
O.~Mees, L.~Hermann, and W.~Burgard, ``What matters in language conditioned
  robotic imitation learning over unstructured data,'' \emph{IEEE Robotics and
  Automation Letters (RA-L)}, vol.~7, no.~4, pp. 11\,205--11\,212, 2022.

\bibitem{lynch2022interactive}
C.~Lynch, A.~Wahid, J.~Tompson, T.~Ding, J.~Betker, R.~Baruch, T.~Armstrong,
  and P.~Florence, ``Interactive language: Talking to robots in real time,''
  \emph{arXiv preprint arXiv:2210.06407}, 2022.

\bibitem{brohan2022rt}
A.~Brohan, N.~Brown, J.~Carbajal, Y.~Chebotar, J.~Dabis, C.~Finn,
  K.~Gopalakrishnan, K.~Hausman, A.~Herzog, J.~Hsu \emph{et~al.}, ``Rt-1:
  Robotics transformer for real-world control at scale,'' \emph{arXiv preprint
  arXiv:2212.06817}, 2022.

\bibitem{ahn2022can}
M.~Ahn, A.~Brohan, N.~Brown, Y.~Chebotar, O.~Cortes, B.~David, C.~Finn,
  K.~Gopalakrishnan, K.~Hausman, A.~Herzog \emph{et~al.}, ``Do as i can, not as
  i say: Grounding language in robotic affordances,'' \emph{arXiv preprint
  arXiv:2204.01691}, 2022.

\bibitem{chen2022open}
B.~Chen, F.~Xia, B.~Ichter, K.~Rao, K.~Gopalakrishnan, M.~S. Ryoo, A.~Stone,
  and D.~Kappler, ``Open-vocabulary queryable scene representations for real
  world planning,'' \emph{arXiv preprint arXiv:2209.09874}, 2022.

\bibitem{liang2022code}
J.~Liang, W.~Huang, F.~Xia, P.~Xu, K.~Hausman, B.~Ichter, P.~Florence, and
  A.~Zeng, ``Code as policies: Language model programs for embodied control,''
  \emph{arXiv preprint arXiv:2209.07753}, 2022.

\bibitem{simeonov2020long}
A.~Simeonov, Y.~Du, B.~Kim, F.~R. Hogan, J.~Tenenbaum, P.~Agrawal, and
  A.~Rodriguez, ``A long horizon planning framework for manipulating rigid
  pointcloud objects,'' \emph{arXiv preprint arXiv:2011.08177}, 2020.

\bibitem{curtis2022long}
A.~Curtis, X.~Fang, L.~P. Kaelbling, T.~Lozano-P{\'e}rez, and C.~R. Garrett,
  ``Long-horizon manipulation of unknown objects via task and motion planning
  with estimated affordances,'' in \emph{2022 International Conference on
  Robotics and Automation (ICRA)}.\hskip 1em plus 0.5em minus 0.4em\relax IEEE,
  2022, pp. 1940--1946.

\bibitem{bobu2022learning}
A.~Bobu, C.~Paxton, W.~Yang, B.~Sundaralingam, Y.-W. Chao, M.~Cakmak, and
  D.~Fox, ``Learning perceptual concepts by bootstrapping from human queries,''
  \emph{IEEE Robotics and Automation Letters}, vol.~7, no.~4, pp.
  11\,260--11\,267, 2022.

\bibitem{yuan2021sornet}
W.~Yuan, C.~Paxton, K.~Desingh, and D.~Fox, ``Sornet: Spatial object-centric
  representations for sequential manipulation,'' in \emph{5th Annual Conference
  on Robot Learning}.\hskip 1em plus 0.5em minus 0.4em\relax PMLR, 2021, pp.
  148--157.

\bibitem{simeonov2022relational}
A.~Simeonov, Y.~Du, L.~Yen-Chen, , A.~Rodriguez, , L.~P. Kaelbling, T.~L.
  Perez, and P.~Agrawal, ``Se(3)-equivariant relational rearrangement with
  neural descriptor fields,'' in \emph{Conference on Robot Learning
  (CoRL)}.\hskip 1em plus 0.5em minus 0.4em\relax PMLR, 2022.

\bibitem{sharma2020relational}
M.~Sharma and O.~Kroemer, ``Relational learning for skill preconditions,''
  \emph{arXiv preprint arXiv:2012.01693}, 2020.

\bibitem{migimatsu2022grounding}
T.~Migimatsu and J.~Bohg, ``Grounding predicates through actions,'' in
  \emph{2022 International Conference on Robotics and Automation (ICRA)}.\hskip
  1em plus 0.5em minus 0.4em\relax IEEE, 2022, pp. 3498--3504.

\bibitem{urain2022se3dif}
J.~Urain, N.~Funk, G.~Chalvatzaki, and J.~Peters, ``Se(3)-diffusionfields:
  Learning smooth cost functions for joint grasp and motion optimization
  through diffusion,'' \emph{https://arxiv.org/pdf/2209.03855.pdf}, 2022.

\bibitem{ajay2022conditional}
A.~Ajay, Y.~Du, A.~Gupta, J.~Tenenbaum, T.~Jaakkola, and P.~Agrawal, ``Is
  conditional generative modeling all you need for decision-making?''
  \emph{arXiv preprint arXiv:2211.15657}, 2022.

\bibitem{vaswani2017attention}
A.~Vaswani, N.~Shazeer, N.~Parmar, J.~Uszkoreit, L.~Jones, A.~N. Gomez,
  {\L}.~Kaiser, and I.~Polosukhin, ``Attention is all you need,'' in
  \emph{Advances in neural information processing systems}, 2017, pp.
  5998--6008.

\bibitem{zhou2019continuity}
Y.~Zhou, C.~Barnes, J.~Lu, J.~Yang, and H.~Li, ``On the continuity of rotation
  representations in neural networks,'' in \emph{Proceedings of the IEEE/CVF
  Conference on Computer Vision and Pattern Recognition}, 2019, pp. 5745--5753.

\bibitem{dhariwal2021diffusion}
P.~Dhariwal and A.~Nichol, ``Diffusion models beat gans on image synthesis,''
  \emph{Advances in Neural Information Processing Systems}, vol.~34, pp.
  8780--8794, 2021.

\bibitem{nair2022learning}
S.~Nair, E.~Mitchell, K.~Chen, S.~Savarese, C.~Finn \emph{et~al.}, ``Learning
  language-conditioned robot behavior from offline data and crowd-sourced
  annotation,'' in \emph{Conference on Robot Learning}.\hskip 1em plus 0.5em
  minus 0.4em\relax PMLR, 2022, pp. 1303--1315.

\bibitem{murali20206}
A.~Murali, A.~Mousavian, C.~Eppner, C.~Paxton, and D.~Fox, ``6-dof grasping for
  target-driven object manipulation in clutter,'' in \emph{2020 IEEE
  International Conference on Robotics and Automation (ICRA)}.\hskip 1em plus
  0.5em minus 0.4em\relax IEEE, 2020, pp. 6232--6238.

\bibitem{lu2020multifingered}
Q.~Lu, M.~Van~der Merwe, B.~Sundaralingam, and T.~Hermans, ``Multifingered
  grasp planning via inference in deep neural networks: Outperforming sampling
  by learning differentiable models,'' \emph{IEEE Robotics \& Automation
  Magazine}, vol.~27, no.~2, pp. 55--65, 2020.

\bibitem{chen2021fastrack}
M.~Chen, S.~L. Herbert, H.~Hu, Y.~Pu, J.~F. Fisac, S.~Bansal, S.~Han, and C.~J.
  Tomlin, ``Fastrack: a modular framework for real-time motion planning and
  guaranteed safe tracking,'' \emph{IEEE Transactions on Automatic Control},
  vol.~66, no.~12, pp. 5861--5876, 2021.

\bibitem{downs2022google}
L.~Downs, A.~Francis, N.~Koenig, B.~Kinman, R.~Hickman, K.~Reymann, T.~B.
  McHugh, and V.~Vanhoucke, ``Google scanned objects: A high-quality dataset of
  3d scanned household items,'' \emph{arXiv preprint arXiv:2204.11918}, 2022.

\bibitem{szot2021habitat}
A.~Szot, A.~Clegg, E.~Undersander, E.~Wijmans, Y.~Zhao, J.~Turner, N.~Maestre,
  M.~Mukadam, D.~Chaplot, O.~Maksymets, A.~Gokaslan, V.~Vondrus, S.~Dharur,
  F.~Meier, W.~Galuba, A.~Chang, Z.~Kira, V.~Koltun, J.~Malik, M.~Savva, and
  D.~Batra, ``Habitat 2.0: Training home assistants to rearrange their
  habitat,'' in \emph{Advances in Neural Information Processing Systems
  (NeurIPS)}, 2021.

\bibitem{calli2015ycb}
B.~Calli, A.~Singh, A.~Walsman, S.~Srinivasa, P.~Abbeel, and A.~M. Dollar,
  ``The ycb object and model set: Towards common benchmarks for manipulation
  research,'' in \emph{2015 international conference on advanced robotics
  (ICAR)}.\hskip 1em plus 0.5em minus 0.4em\relax IEEE, 2015, pp. 510--517.

\bibitem{coumans2017pybullet}
E.~Coumans and Y.~Bai, ``Pybullet, a python module for physics simulation in
  robotics, games and machine learning,'' 2017.

\bibitem{morrical2021nvisii}
N.~Morrical, J.~Tremblay, Y.~Lin, S.~Tyree, S.~Birchfield, V.~Pascucci, and
  I.~Wald, ``Nvisii: A scriptable tool for photorealistic image generation,''
  \emph{arXiv preprint arXiv:2105.13962}, 2021.

\bibitem{wu2022targf}
M.~Wu, F.~Zhong, Y.~Xia, and H.~Dong, ``Targf: Learning target gradient field
  for object rearrangement,'' \emph{arXiv preprint arXiv:2209.00853}, 2022.

\bibitem{rusu20113d}
R.~B. Rusu and S.~Cousins, ``3d is here: Point cloud library (pcl),'' in
  \emph{2011 IEEE international conference on robotics and automation}.\hskip
  1em plus 0.5em minus 0.4em\relax IEEE, 2011, pp. 1--4.

\bibitem{ten2018using}
A.~Ten~Pas and R.~Platt, ``Using geometry to detect grasp poses in 3d point
  clouds,'' in \emph{Robotics research}.\hskip 1em plus 0.5em minus 0.4em\relax
  Springer, 2018, pp. 307--324.

\bibitem{kent2018adaptive}
D.~Kent and R.~Toris, ``Adaptive autonomous grasp selection via pairwise
  ranking,'' in \emph{2018 IEEE/RSJ International Conference on Intelligent
  Robots and Systems (IROS)}.\hskip 1em plus 0.5em minus 0.4em\relax IEEE,
  2018, pp. 2971--2976.

\bibitem{sundermeyer2021contact}
M.~Sundermeyer, A.~Mousavian, R.~Triebel, and D.~Fox, ``Contact-graspnet:
  Efficient 6-dof grasp generation in cluttered scenes,'' in \emph{2021 IEEE
  International Conference on Robotics and Automation (ICRA)}.\hskip 1em plus
  0.5em minus 0.4em\relax IEEE, 2021, pp. 13\,438--13\,444.

\bibitem{kuffner2000rrt}
J.~J. Kuffner and S.~M. LaValle, ``{RRT}-connect: An efficient approach to
  single-query path planning,'' in \emph{Proceedings 2000 ICRA. Millennium
  Conference. IEEE International Conference on Robotics and Automation.
  Symposia Proceedings (Cat. No. 00CH37065)}, vol.~2.\hskip 1em plus 0.5em
  minus 0.4em\relax IEEE, 2000, pp. 995--1001.

\bibitem{zhao2020towards}
J.~Zhao, D.~Troniak, and O.~Kroemer, ``Towards robotic assembly by predicting
  robust, precise and task-oriented grasps,'' in \emph{CoRL}, 2022.

\bibitem{reimers-2019-sentence-bert}
\BIBentryALTinterwordspacing
N.~Reimers and I.~Gurevych, ``Sentence-bert: Sentence embeddings using siamese
  bert-networks,'' in \emph{Proceedings of the 2019 Conference on Empirical
  Methods in Natural Language Processing}.\hskip 1em plus 0.5em minus
  0.4em\relax Association for Computational Linguistics, 11 2019. [Online].
  Available: \url{https://arxiv.org/abs/1908.10084}
\BIBentrySTDinterwordspacing

\end{thebibliography}

\renewcommand{\thesection}{A.\arabic{section}}
\renewcommand{\thefigure}{A\arabic{figure}}
\renewcommand{\thetable}{A\arabic{table}}

\setcounter{section}{0}
\setcounter{figure}{0}
\setcounter{table}{0}

\renewcommand{\theequation}{A\arabic{equation}}
\setcounter{equation}{0}

\clearpage
\appendix



\subsection{Towards Natural Language}
\label{sec:sentence-model}
In the main experiments, we evaluate the model that encodes word tokens individually. Here, we want to see if our framework can deal with more natural language without predefined vocabularies and at different levels of abstraction (e.g., ``make a line'' vs ``make a short line at the bottom of the table'') by combining with a pretrained language model.

\subsubsection{Language Data}
We procedurally generate sentences to train the model. At test time, we leverage sentence embeddings to generalize to new sentences. Given the word tokens in sentences, we first enumerate all possible combinations of the tokens. Note in our definition each token corresponds to a type and a value. For example, [(shape, circle), (size, small), (x\_position, center), (y\_position, middle), (rotation, south)] can create combinations such as [(shape, circle)], [(size, small), (x\_position, center)], [(shape, circle), (x\_position, center), (y\_position, middle)].  In total, there are 669 unique combinations of word tokens. We then manually created templates for each combination of token types. For example, ``build a [shape]", ``put the objects into [shape]", and ``build a [size] [shape] on the [x\_position] of the table facing [rotation]". Combining the enumerated combinations and sentence templates, we generated 3345 unique sentences. We then map all the sentences to pretrained sentence embeddings using a sentence transformer~\cite{reimers-2019-sentence-bert}, specifically the \lstinline|all-MiniLM-L6-v2| model. Note that although the generated sentences could have grammar errors, the pretrained sentence transformer should be robust to those errors.

\subsubsection{Model and Training} 
We modify the diffusion model to take in a single sentence embedding instead of the embeddings of individual word tokens. The sentence embedding from the pretrained model has a dimension of 384. We use a linear layer to map it to the same dimension as the point cloud embeddings. During training, we randomly sample a combination of the word tokens and map them to predefined sentence embeddings. Therefore, the model should be able to map ambiguous sentences (e.g., build a structure in the middle) to different structures (e.g., a line, a circle, and etc).

\subsubsection{Inference}
\begin{figure}[tbh!]
\includegraphics[width=\columnwidth]{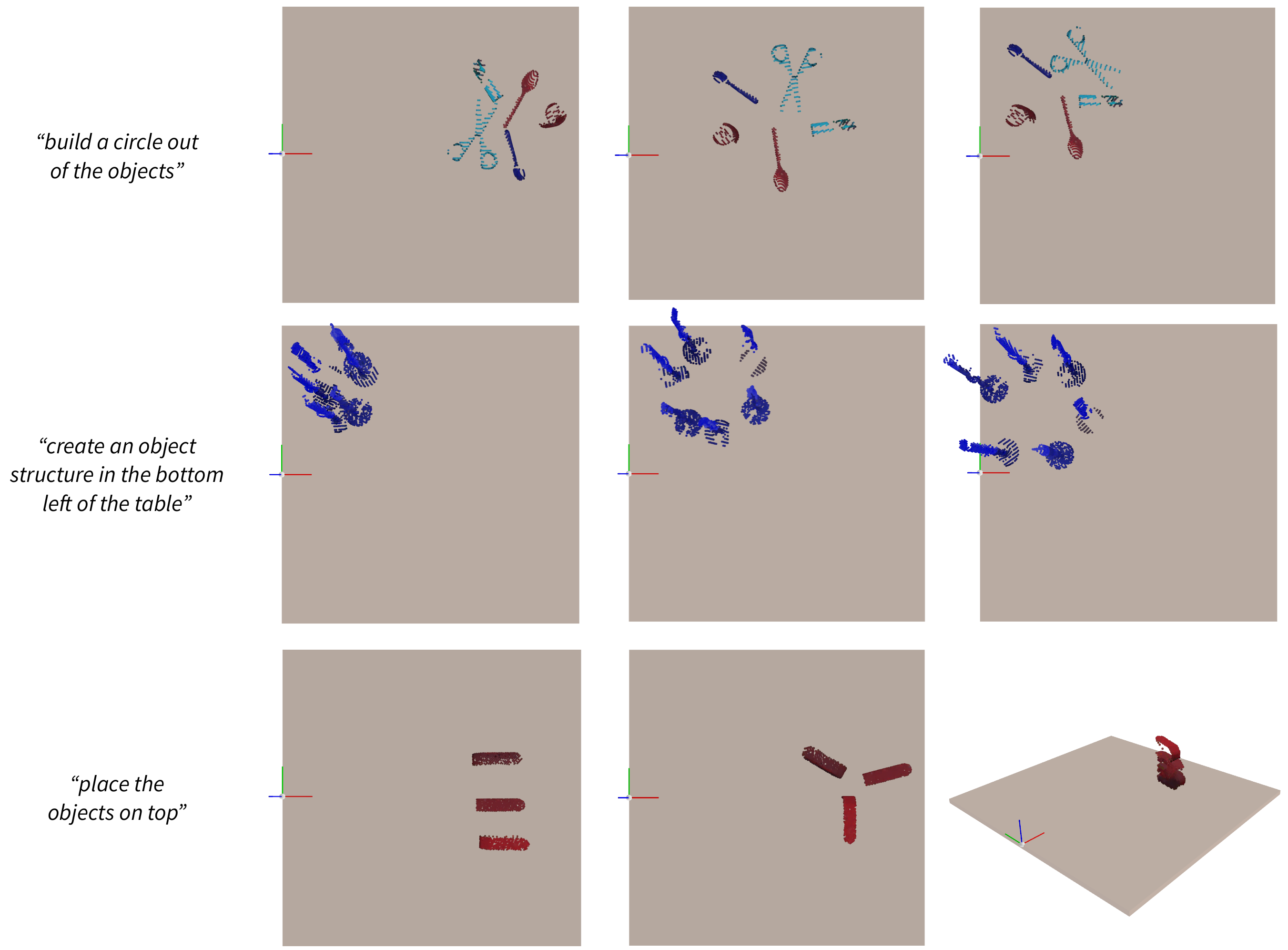}
\caption{Each row shows three sampled goals from the Diffusion model that is trained with procedurally generated sentences. The model generalizes to input sentences never seen during training by using a pretrained sentence encoder. Because the model is trained on language inputs at different levels of abstractions, the generated samples correspond strongly to structure properties specified in the sentences and at the same time show variations for the properties that are left unspecified.}
\label{fig:sentence-model-example}
\end{figure}

We evaluate the performance of the model using fully-specified sentences. As shown in Table~\ref{tab:sentence-model}, we see within 5\% decrease in performance across structure types compared to the original model. Leveraging the generalization capability of the pretrained language model, the model should also be able to generalize to sentences not in the rearrangement training data at test time. We show some qualitative examples in Figure~\ref{fig:sentence-model-example}.

\subsubsection{Using the Large Language Model to Parse Natural Language}
We also explore the prompting techniques \cite{liang2022code,ahn2022can} to map natural language to word tokens that our original model knows.  Examples in Listing~\ref{lst:label} shows that this is also a promising direction to bridge the gap between natural language and domain-specific commands.

\def\graycolor{\color{gray}}
\def\purplecolor{\color{purple}}
\def\greencolor{\color{LimeGreen}}
\begin{lstlisting}[frame=single,escapechar=@,float,floatplacement=H,caption={Example GPT-3 Prompt to map natural language descriptions of target structures to word tokens that our model consumes. Prompt context is in \textcolor{gray}{gray}, input language descriptions are \textcolor{LimeGreen}{green}, and generated word tokens are \textcolor{purple}{purple}.},label={lst:label}]
@\aftergroup\graycolor@Map each natural language instruction about an 
object arrangement to available shape, size, and 
location. If any element is not specified, 
return NONE.

Available shapes: circle, line, table setting, tower
Available size: small, medium, large
Available location: top, middle, bottom

Example: rearrange the objects into a small circle.
Output: circle, small, NONE

@\aftergroup\greencolor@Example: put the objects far apart and as a line 
on top of the table.
Output: @\aftergroup\purplecolor@line, large, top

@\aftergroup\greencolor@Example: there is a chair near the bottom edge 
of the table. Arrange the dishes near the chair.
Output: @\aftergroup\purplecolor@table setting, NONE, bottom

@\aftergroup\greencolor@Example: There is a teapot on the top right corner 
of the table. Place the teacups surrounding it.
Output: @\aftergroup\purplecolor@circle, small, top
\end{lstlisting}

\begin{table}[tbh!]
  \centering
  \caption{Comparing the success rate \% in simulation for the two types of language inputs. We observe a small drop in performance when we use a single sentence embedding comparing to individually embed each word tokens for each input sentence.}
  \begin{tabular}{lcccc}
    \toprule
    Language & Circle& Tower& Line & Table Setting \\
    \midrule
     Word Tokens & 65 & 72 & 72 & 66\\
     Sentence & 63 & 70 & 67 & 67\\
    \bottomrule
  \end{tabular}
  \label{tab:sentence-model}
\end{table}

\begin{table*}[t]
  \centering
  \caption{The percentage of failure cases due to different types of errors for different models averaged over all structure types.}
  \begin{tabular}{lcccc}
    \toprule
    Model & Intersection & Abnormal Velocity& Not Upright & Semantic Failure \\
    \midrule
    StructDiffusion w/o Collision Discriminator & 17.33& 14.36& 5.20& 5.94\\
    StructDiffusion & 6.44& 17.57& 4.46& 3.71\\
    CVAE  & 28.71& 14.11& 5.20& 5.69\\
    CVAE w/ Collision Discriminator & 20.05& 12.62& 3.22& 3.47\\
    Opt w/ CVAE Prior & 17.33& 13.12& 5.69& 6.93\\
    \bottomrule
  \end{tabular}
  \label{tab:failure-mode-methods}
\end{table*}

\begin{table*}[t]
  \centering
  \caption{The percentage of failure cases due to different types of errors for StructDiffusion.}
  \begin{tabular}{lcccc}
    \toprule
    Structure Type & Intersection & Abnormal Velocity& Not Upright & Semantic Failure \\
    \midrule
    Circle & 4.95 & 24.75& 4.95&0.00\\
    Tower & 12.87& 8.91& 4.95& 8.91\\
    Line  &  5.94& 17.82 &4.95&0.0\\
    Table Setting & 1.98& 18.81 &2.97 &5.94  \\
    \bottomrule
  \end{tabular}
  \label{tab:failure-mode-structures}
\end{table*}

\subsection{Classifier-Based Collision Guidance}
\label{sec:classifier-guidance}
In addition to utilizing a separate collision discriminator, we may directly integrate a classifier into the underlying sampling procedure of the diffusion model by using classifier based guidance. We present results in Table \ref{tbl:collision}.  We find that integrating a noisy collision discriminator with sampling can improve performance for towers and table settings but leads to limited gains for other structures.

\subsection{Failure Mode Analysis}
\label{sec:failure-mode-analysis}

We analyze the failure modes for our model based on the simulation experiment. In Table~\ref{tab:failure-mode-methods}, we first compare the number of failure cases in all simulation experiments for different methods. We observe that \structdiffusion{}, which combines a collision discriminator and a diffusion model, drastically reduces errors due to intersections between objects.  Comparing CVAE and Optimization with CVAE prior, we also observe that optimizing the goal poses with a learned discriminator can reduce the intersection errors but at the expense of generating more semantically incorrect structures.  In Table~\ref{tab:failure-mode-structures}, we further break down the failure modes for our method on different structure types.  Building circles tend to have a higher failure rate due to abnormal velocity because large circles may have objects occasionally placed outside of the table. In comparison, towers suffer less from this issue because objects are more closely packed.  Reasoning about the environment (e.g., the size of the table) is a necessary next step to address this issue. 

\subsection{Real World Results}
\label{sec:extra-real-world}
\begin{figure*}[hbt!]
\includegraphics[width=\textwidth]{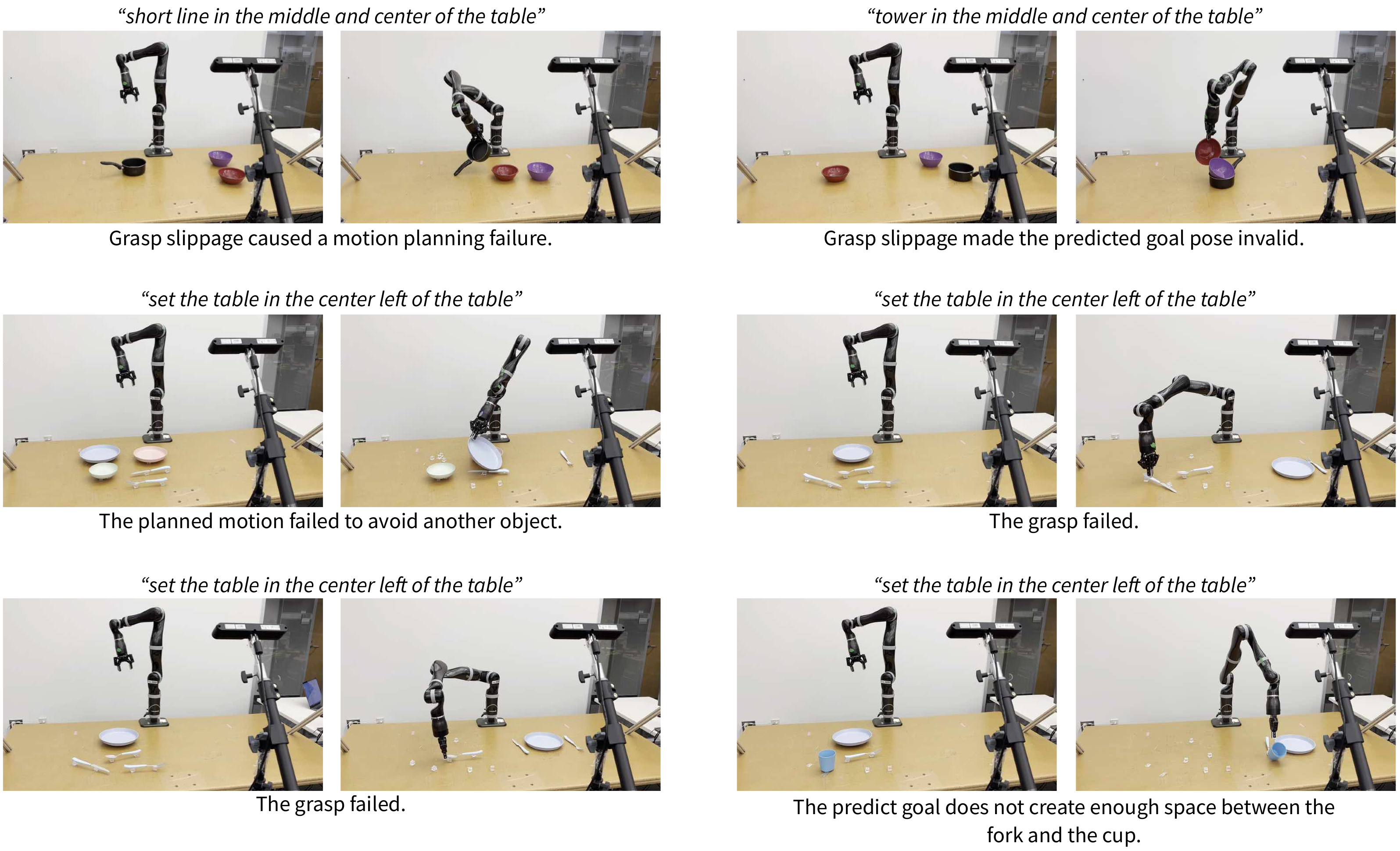}
\caption{Real-world failure cases from our robot experiments. A minority of failures come from StructDiffusion; the majority from grasping, planning, and execution failures. This points to future work integrating StructDiffusion with task and motion planning or with }
\label{fig:real-world-failures}
\end{figure*}

\begin{figure*}[hbt!]
\includegraphics[width=\textwidth]{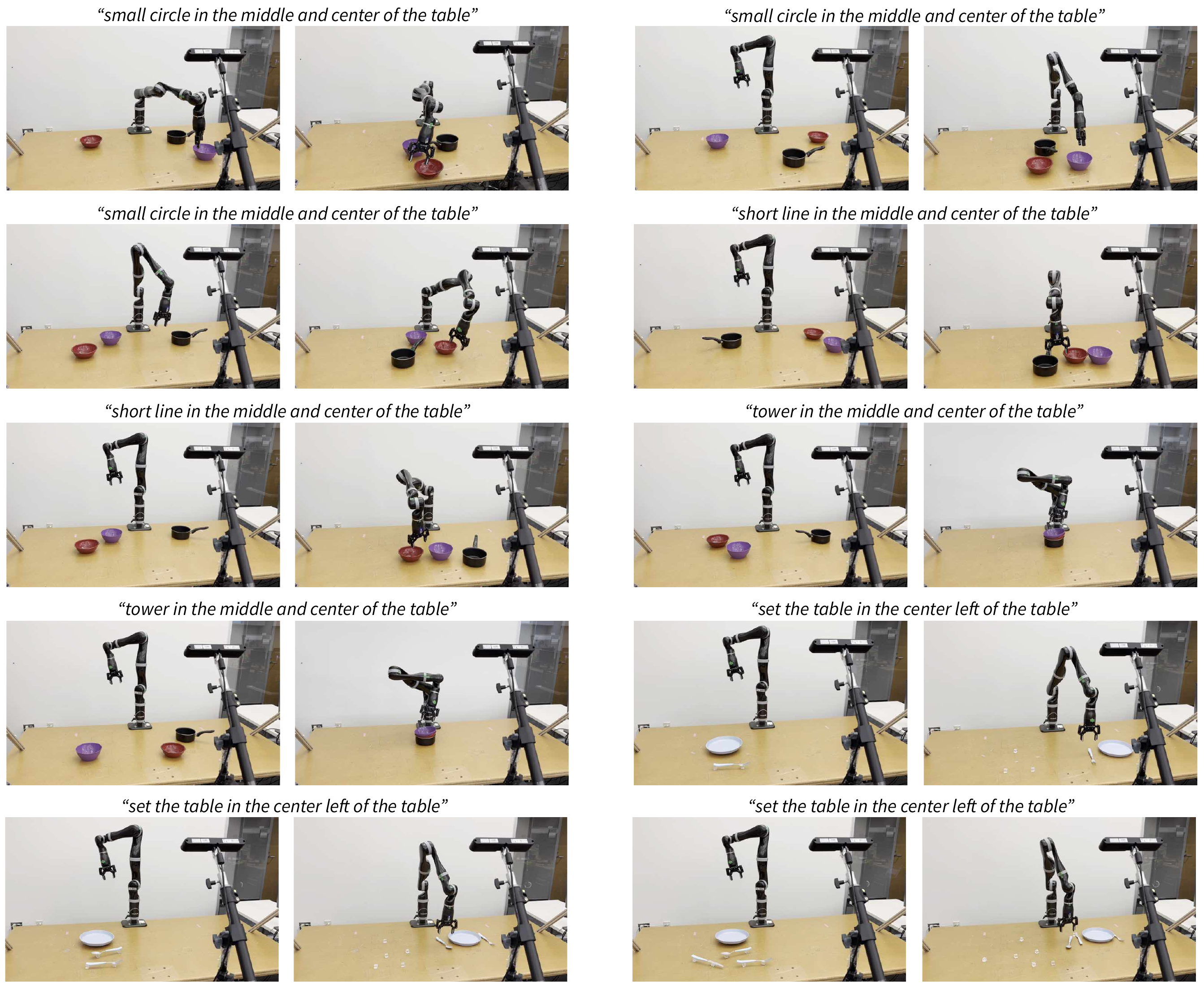}
\caption{Real-world success cases from our robot experiments. Out of 20 experiments, we saw 11 successes. Most failures were due to grasping and motion planning, not due to issues with our method. All experiments were performed on unseen objects, and StructDiffusion was not trained on any real-world training data.}
\label{fig:real-world-successes}
\end{figure*}

Here we show results from each of the real-world experiments in Table~\ref{tab:robot_result}. Figure~\ref{fig:real-world-successes} shows some additional success examples. Success cases show how StructDiffusion allows us to rearrange previously-unseen objects that did not appear in the training data.
Figure~\ref{fig:real-world-failures} includes examples of cases where our method fails. In summary, we see that most of our failures were due to simple motion planning or grasping issues; these could be solved in the future via better integration with task and motion planning algorithms, and with reactive replanning.

\begin{table}[t]
\centering
\caption{Effect of Collision Classifier Guidance on success rate. This table shows the effect noisy classifier guidance on performance on each of the provided tasks. We do not see clear advantages to using the collision classifier.}
\begin{tabular}{lccccc}
    \toprule
      & \multicolumn{5}{c}{\bf Guidance Weight} \\
      \cmidrule(lr){2-6}
      {\bf Task} & 0 & 1 & 2 & 4 & 8 \\
      \midrule
      Circle & 65 & 60 & 58 &  58  &  61\\
      Tower & 72 & 76 & 74 & 78 &  73\\
      Line & 72 & 73 & 69 & 73 & 69 \\
      Table Setting & 66 & 70  & 73 & 66 & 70 \\
    \bottomrule
\end{tabular}
\label{tbl:collision}
\end{table}

\end{document}